\documentclass{cta-author}

\usepackage{algorithmic}
\usepackage{graphicx}
\usepackage{textcomp}
\usepackage{wrapfig}
\usepackage{subfigure}
\usepackage{multirow}
\usepackage{threeparttable}
\usepackage{booktabs}
\usepackage{hhline}
\usepackage{ulem}
\usepackage{hyperref}
\hypersetup{
	colorlinks=true,
	linkcolor=blue,
	filecolor=blue,      
	urlcolor=blue,
	citecolor=cyan,
}

\usepackage{array}
\usepackage{colortbl}

{}
{}
{}

\begin{document}
	
	
	\title{Exploiting Robust Unsupervised Video Person Re-identification}
	
	\author{\au{Xianghao Zang$^{1}$}, \au{Ge Li$^{1}$}, \au{Wei Gao$^{1\corr}$}, \au{Xiujun Shu$^{2}$}}
	
	\address{\add{1}{School of Electronic and Computer Engineering, Peking University, Shenzhen 518055, China.}
		\add{2}{Peng Cheng Laboratory, Shenzhen 518034, China.}
		\email{gaowei262@pku.edu.cn}}
	
	\begin{abstract}
		Unsupervised video person re-identification (reID) methods usually depend on global-level features. And many supervised reID methods employed local-level features and achieved significant performance improvements. However, applying local-level features to unsupervised methods may introduce an unstable performance. To improve the performance stability for unsupervised video reID, this paper introduces a general scheme fusing part models and unsupervised learning. In this scheme, the global-level feature is divided into equal local-level feature. A local-aware module is employed to explore the poentials of local-level feature for unsupervised learning. A global-aware module is proposed to overcome the disadvantages of local-level features. Features from these two modules are fused to form a robust feature representation for each input image. This feature representation has the advantages of local-level feature without suffering from its disadvantages. Comprehensive experiments are conducted on three benchmarks, including PRID2011, iLIDS-VID, and DukeMTMC-VideoReID, and the results demonstrate that the proposed approach achieves state-of-the-art performance. Extensive ablation studies demonstrate the effectiveness and robustness of proposed scheme, local-aware module and global-aware module. The code and generated features are available at \href{https://github.com/deropty/uPMnet}{https://github.com/deropty/uPMnet}.
	\end{abstract}
	
	\maketitle
	
	\section{Introduction} \label{introduction}
	Person re-identification (reID) aims to find the correct person from gallery for the person of interest in query within non-overlapping cameras, which is an important practical application in surveillance camera network \cite{RN528} \cite{RN527} \cite{RN529}. In recent years, video-based reID has gained increasing attention because video sequences can provide rich temporal and spatial information for a specific person identity \cite{RN424} \cite{RN425} \cite{RN418}. Meanwhile, with development of convolution neural networks (CNN), supervised video reID has been improved gradually. However, as the number of cameras increases, practical applications are largely limited due to the growing cost of labeling work. Consequently, this situation motivates researchers to develop unsupervised methods.
	
	In the context of unsupervised learning, approaches for reID can be roughly divided into four categories: 1) transfer learning-based methods \cite{RN495} \cite{RN372} \cite{RN353}; 2) one-shot learning-based methods \cite{RN390} \cite{RN107} \cite{RN95}; 3) clustering-based methods \cite{RN360} \cite{RN247} \cite{RN118}; 4) tracklet-based association learning \cite{RN97} \cite{RN87} \cite{RN369} \cite{RN389}. These methods above employed global-level features to achieve unsupervised learning. However, more detailed feature representation often introduces remarkable performance improvement. For example, supervised methods for reID task employed local-level features \cite{RN191} \cite{RN229}, texture semantics \cite{RN452} \cite{RN477}, skeleton point information \cite{RN472} \cite{RN108} \cite{RN291}, etc., and achieve a better performance. 
		
	\begin{figure}[!h]
		\centering{\includegraphics[width=0.5\textwidth]{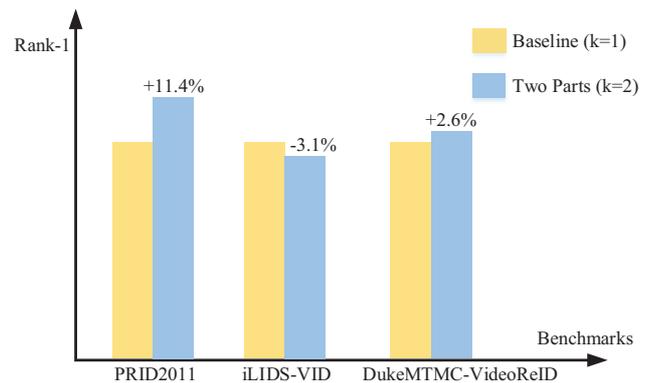}}
		\caption{Directly applying part models to unsupervised video reID has an unstable performance (k is the number of parts). The Baseline method only employs a global-level feature, \textit{i.e.,} k=1. `+/-' means the percentage increase or decrease relative to the baseline method. } \label{motivation_new}
	\end{figure}
	
	It is well known that part models-based methods are much common for the reID task in a supervised setting, even overstudied \cite{RN191} \cite{RN456} \cite{RN231} \cite{RN472} \cite{RN448} \cite{RN229}. These methods improved performance significantly in the context of supervised learning. For unsupervised learning, the performance of part models remains to be explored. Thus, the unsupervised method DAL \cite{RN187} is employed directly to train each local-level feature individually. MobileNet \cite{RN228} is used as the backbone for rapid deployment. For different benchmarks, the performances are uncertain, as illustrated in Fig. \ref{motivation_new}. Significantly, there is an apparent performance drop for the benchmark iLIDS-VID \cite{RN385}.
	
\begin{figure}[!h]
		\centering{\includegraphics[width=0.48\textwidth]{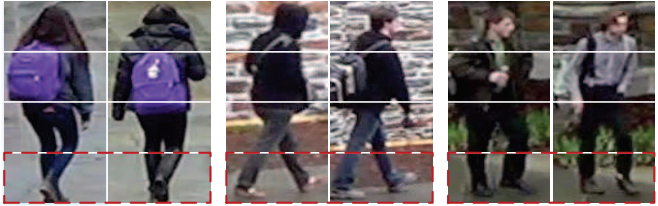}}
		\caption{The same body parts of different persons usually have a more similar appearance than their holistic image. For example, each image pair above has similar foot appearances.} \label{motivation_contrastive}
	\end{figure}

	The possible reason for these uncertain performances is below. For the reID task, the same body parts of different persons usually have a more similar appearance than their holistic image \cite{RN98}, as illustrated in Fig. \ref{motivation_contrastive}. These similar appearances make it hard to distinguish the same body parts of different persons, which introduces a negative effect. On the other hand, the dimension of concatenated local-level features is several times the global one, which presents a richer feature representation and has a positive effect. These contrary effects of part models on unsupervised learning lead to uncertain performances.
	
	To improve the performance stability, this paper introduces a robust part models-based scheme for unsupervised video reID, as illustrated in Fig. \ref{scheme}. Global-level feature is divided into equal local-level features. 
	The global-level feature and local-level features are collected and delivered to local/global-aware module. The local-aware module is used to explore the potentials of local-level features for unsupervised learning. The global-aware module is introduced to overcome the disadvantages of local-level features. The global-aware module increases the robustness of proposed scheme in various challenging situations and has a complementary performance with the local-aware module. The proposed scheme is trained twice where the local-aware module and the global-aware module are incorporated, respectively. The features from these two modules are fused to form a rich and robust feature representation. This feature representation has the superiority of part models and discards its shortcomings. Through extensive experiments, proposed scheme achieves state-of-the-art performance on three benchmarks, \textit{i.e.,} PRID2011 \cite{RN386}, iLIDS-VID \cite{RN385}, and DukeMTMC-VideoReID \cite{RN170} \cite{RN391}. The ablation study also demonstrates the effectiveness of proposed scheme and modules. 
	
	The main contributions of this paper can be summarized as follows:
	\begin{itemize}
		\item This paper introduces a robust part models-based scheme for unsupervised video reID, which makes it easier to explore the different effects of part models on unsupervised learning.
		\item A global-aware module is proposed to improve the robustness of part models. The features from the local-aware module and global-aware module are fused to form a robust feature representation for a better unsupervised video reID.
		\item Experimental results demonstrate that proposed method achieves significant performance gains. Specifically, the proposed method achieves Rank-1 improvements of 6.7\% on PRID2011, 5.8\% on iLIDS-VID, and 0.8\% on DukeMTMC-VideoReID, compared to other state-of-the-art unsupervised methods.
	\end{itemize}
	
	The rest of this paper is organized as follows. The related works are reviewed and analyzed in Section \ref{related work}, and then proposed method is introduced in Section \ref{method}. Experimental results and analysis are presented in Section \ref{Experiments}, and Section \ref{conlusion} concludes this paper.
	
	\section{Related Work}\label{related work}
	Most video reID methods are formulated in a supervised manner, which hinders their deployment in real-world applications.  
	
	\subsection{Unsupervised Video Person Re-identification}
	Unsupervised video reID has attracted increasing research interest recently \cite{RN96} \cite{RN370} \cite{RN309} \cite{RN357}. These methods can be roughly divided into four categories as explained in Section \ref{introduction}. 
	
	\paragraph{Transfer Learning}
	Transfer learning-based methods, also called domain adaption-based methods, is an important branch of unsupervised approaches, aiming to transfer the source domain-trained model to the target domain \cite{RN104} \cite{RN221}. Huang \textit{et al.} \cite{RN359} introduced a domain adaptive attention model (DAAM) to separate the feature map into domain-shared feature map and domain-specific feature map. These two features were used to improve the performance in target domain and alleviate the negative effects introduced by domain divergence, respectively. However, these transfer learning-based methods require one labeled source dataset, which still needs a lot of labeling work.
	
	\paragraph{One-shot Learning}
	One-shot learning-based methods \cite{RN107} \cite{RN95} often assume the prior spatio-temporal topology knowledge could be used to achieve person identities (labels). Specifically, Wu \textit{et al.} \cite{RN391} proposed a dynamic sampling strategy, which started with easy and reliable unlabeled samples and incorporated diverse tracklets, to update their model. Ye \textit{et al.} \cite{RN107} designed an anchor embedding method with regularized affine hull and a manifold smoothing term for this task. Ye \textit{et al.} \cite{RN95} introduced a dynamic graph matching framework to estimate cross-camera labels and used learned metrics to update camera graphs dynamically. However, these methods still require one labeled tracklet per identity for model initialization, which only partially reduces the annotation workload.
	
	\paragraph{Clustering Analysis}
	The clustering-based method is a long-standing paradigm for unsupervised learning. With the surge of deep CNN, recent studies have attempted to optimize clustering analysis and representation learning jointly \cite{RN247} \cite{RN118}. Lin \textit{et al.} \cite{RN247} proposed a bottom-up clustering approach that jointly optimized a CNN model and the relationship among the individual samples. Ding \textit{et al.} \cite{RN118} introduced a dispersion-based criterion to evaluate the quality of the automatically generated clusters, which showed the importance of cluster validity quantification. However, only focusing on cluster-level repelling and merging neglects the latent variational information among the individual samples.
	
	\paragraph{Tracklet Association Learning}
	Recently, some tracklet-based methods \cite{RN87} \cite{RN369} \cite{RN389} have been proposed to achieve unsupervised video reID. Chen \textit{et al.} \cite{RN87} and Li \textit{et al.} \cite{RN369} explored end-to-end learning architectures to associate within-camera and cross-camera tracklets by optimizing specifically tailored objective functions. Wu \textit{et al.} \cite{RN389} designed multiple unsupervised learning objectives including tracklet frame coherence, tracklet neighborhood compactness, and tracklet cluster structure in a unified formulation for both image-based and video-based unsupervised reID. However, these methods extract global-level features as compact representation, which neglects the fine-grained part information (\textit{e.g.,} head, torso) and constrains the model performance.
	
	\subsection{Part-based Person Re-identification}
	Employing local-level features for pedestrian image description offers fine-grained information and has been verified as beneficial for person retrieval in recent studies. These part-based methods fall into three categories: 1) using pose estimator to extract a pose map \cite{RN265}; 2) leveraging saliency map to implicitly estimate body part \cite{RN266}; 3) exploiting local-level features introduced by horizontal stripes of multiple scales \cite{RN191} \cite{RN318} \cite{RN394} \cite{RN314} \cite{RN98}. 
	
	\paragraph{Pose-guided Learning}
	The pose landmarks can indicate different body parts \cite{RN145} \cite{RN472} \cite{RN108}. Miao \textit{et al.} \cite{RN291} introduced pose-guided feature alignment (PGFA) to disentangle the useful information from the occlusion noise. The pose-guided global-level feature and partial features were used to measure the distances of possible image pairs. Gao \textit{et al.} \cite{RN533} employed a pose estimator to perform spatial and temporal alignment and selected image sets with the same pose and high quality for video reID. However, these methods above often depend on a pre-trained pose estimator which introduces inevitable bias due to the differences between the pre-trained dataset and reID dataset.
	
	\begin{figure*}[!h]
		\centering{\includegraphics[width=1\textwidth]{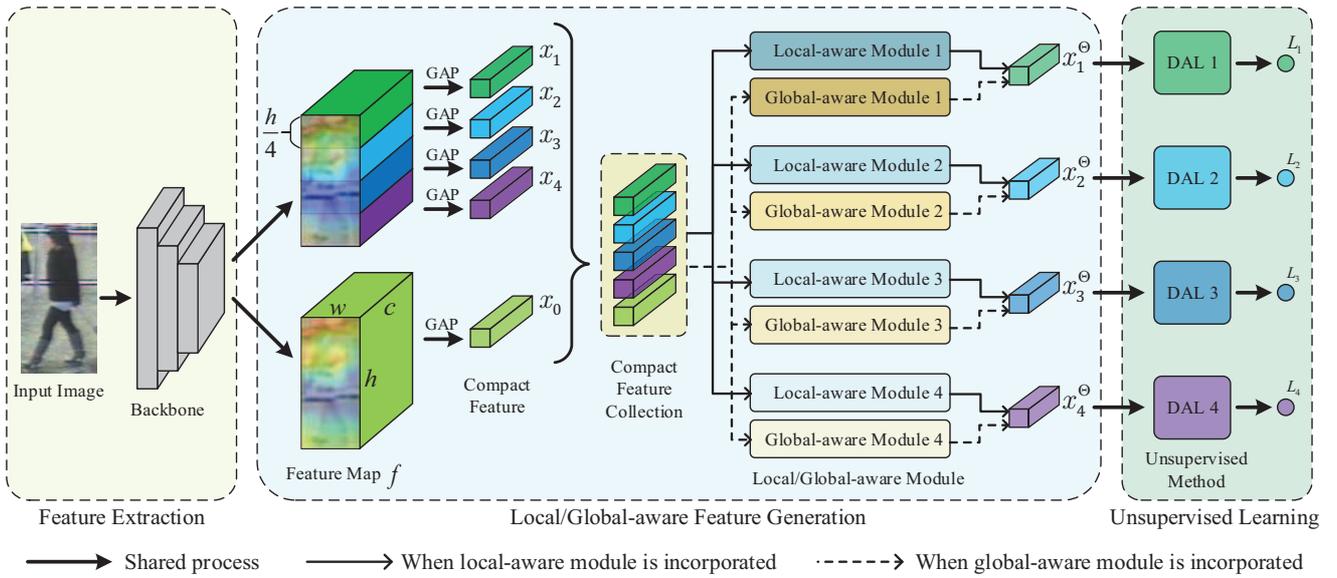}}
		\caption{Overview of proposed scheme, which mainly consists of three parts: 1) extracting feature map from the input person image; 2) generating the local/global-aware features from the feature map; 3) achieving sophisticated unsupervised learning by employing unsupervised modules. The partition scale $k$ is set to 4 in this diagram. \label{scheme}} 
	\end{figure*}
	
	\paragraph{Saliency Learning}
	Salient regions of person images provide valuable information for reID task \cite{RN484} \cite{RN216}. Zhao \textit{et al.} \cite{RN100} employed human salience to extract salient patches and utilized patch matching with adjacency constraint to build dense correspondence for different image pairs in an unsupervised manner. He \textit{et al.} \cite{RN478} utilized the saliency heatmap to generate the adaptive weights for different image parts and achieved guided adaptive spatial matching (GASM) for reID task. However, these methods above depend on the off-the-shelf saliency detector which also introduces inevitable bias.
	
	\paragraph{Stripe/Part-related Learning}
	Stripe/Part-related learning employs a simple strategy and divides the feature map into several equal parts \cite{RN191} \cite{RN314}. Sun \textit{et al.} \cite{RN191} employed this uniform partition strategy and assembled the part-informed features into a convolutional descriptor, which significantly improved the performance for supervised reID. Considering the relationships between one body part and other body parts, Park and Ham \cite{RN314} presented a relation network that makes each local-level feature more discriminative. These methods above employed the person identity to supervise the model to learn each part individually. Then these methods assembled local-level features for the test. This strategy gave each feature discriminative information and introduced a remarkable performance improvement. However, applying part models to unsupervised video reID may introduce an unstable performance.

	\section{Method}\label{method}
	The proposed scheme is illustrated in Fig. \ref{scheme}, which is an unsupervised Part Models-based network (uPMnet) for the video reID task. 

	\subsection{Preliminaries}
	Given a large quantity of video data captured by disjoint surveillance cameras $\{ C_1,C_2,\cdots \}$, each camera $C$ contains a varying number of tracklets $T$, as $C = \{ T_1,T_2,\cdots \}$. Each tracklet $T$ consists of multiple person images $I$, as $T = \{ I_1,I_2,\cdots \}$. The tracklet information (tracklet ID) and camera information (camera ID) of each image are automatically labeled during annotation. Thus it is assumed that these pieces of information are available for the unsupervised video reID. 

	The function of each part in Fig. \ref{scheme} is explained below. In the first part of proposed scheme, \textit{i.e., feature extraction}, CNN is employed to extract an initial feature map $f$ with a size of $h \times w \times c$ ($h, w, c$ are height, width, and the number of channels, respectively) from a person image. In the second part, \textit{i.e., local/global-aware feature generation}, feature map $f$ is divided equally into $k$ horizontal stripes, which follows the operations in \cite{RN191} \cite{RN314} \cite{RN448}. Each stripe has a smaller receptive field than the global-level feature $f$. These stripes are local-level features and are called part models. These stripes and original feature map are followed by global average pooling (GAP) and normalization to generate the local-level features $\{x_i\}_{i=1}^k$ and global-level feature $x_0$. Each of these features is regarded as a \textit{compact feature} with a size of $1 \times 1 \times c$. All the above features are collected and delivered to aware modules which are designed to deal with the local-level and global-level features and output local/global-aware features for the following part. The local-aware module and global-aware module are introduced in Section \ref{local-aware} and Section \ref{global-aware}. In the third part of this scheme, \textit{i.e., unsupervised learning}, $k$ modules are utilized to achieve unsupervised learning. These modules have the same operation but different network parameters. Each module independently handles the corresponding feature. Section \ref{unsupervised learning} introduces the third part of this scheme. And Section \ref{model training and testing} presents the details of model training and testing. 

	In supervised methods \cite{RN318} \cite{RN394}, the global max pooling (GMP) is employed to generate the \textit{compact feature}. The GMP operation can aggregate the feature from the most discriminative parts in person image. In this paper, global average pooling (GAP) is empirically used for unsupervised learning, and the GAP operation introduces an undiscriminating feature representation covering the whole image. The experimental details will be discussed in Section \ref{GAP vs. GMP}.
	
	\subsection{Local-aware Module} \label{local-aware}
	In the second part of proposed scheme, the aware module can be local-aware module or global-aware module. The local-aware module is introduced first as,
	\begin{equation}
		x_i^l = x_i, ~~~~i = (1,\cdots, k), \label{local-aware module}
	\end{equation}
	where only the local-level features $\{x_i\}_{i=1}^k$ are used. These local-level features can introduce significant performance improvement for supervised reID \cite{RN191}. Here the potentials of local-level features on unsupervised learning are explored. Therefore, only the local-level features are employed. The Eq. \ref{local-aware module} is denoted as local-aware module. The compact local-level features $\{x_i\}_{i=1}^k$ are regarded as local-aware features $\{x_i^l\}_{i=1}^k$. 
	
	For unsupervised problems, the performance of local-aware feature is uncertain. First, the concatenated local-aware feature is $k$ times the original global-level feature because each local-aware feature and global-level feature have the same dimension. The feature with a larger dimension often introduces better performance \cite{RN191}. Second, each part has a smaller receptive field. The difference between the same body part of different persons is much smaller than the difference between the different persons. Distinguishing these local-level features is much harder than distinguishing the global-level features \cite{RN98}. Employing part models may be harmful to unsupervised learning. Therefore, compared to global-level features, the part models introduce positive and negative effects. The local-aware module is proposed to explore the pros and cons of part models for the unsupervised video reID task.
	
	\subsection{Global-aware Module} \label{global-aware}
\begin{figure}[!h]
		\centering{\includegraphics[width=0.48\textwidth]{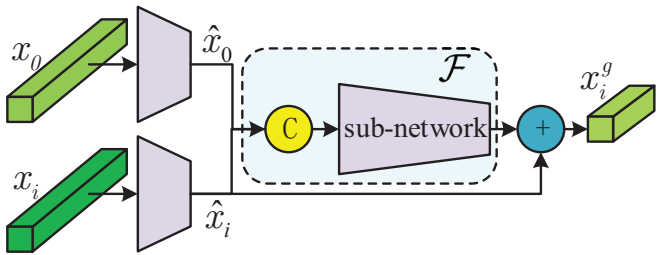}}
		\caption{Illustration of $i^\text{th} (i=1,\cdots,k)$ global-aware module. There are $k$ global-aware modules in our framework. `C' and `+' represent the concatenation operation and element-wise addition operation, respectively. \label{global-aware-module}} 
	\end{figure}

	The global-aware module is introduced in this section, and the motivation is explained first. When many background clutter and occlusions appear on an image, the same part of different images may give a more similar feature representation than their holistic images. In this situation, the negative effect of part models is amplified. A global-aware module is proposed to optimize this issue, as illustrated in Fig. \ref{global-aware-module}. Concretely, a $1\times 1$ convolutional layer is added to each local-level feature $x_i (i=1,\cdots, k)$ and global one $x_0$ to generate feature vectors $\{\hat{x}_i\}_{i=0}^k$ with a size of $1 \times 1 \times c'$. Then $\hat{x}_0$ is concatenated to the end of each local-level feature $\hat{x}_i (i=1,\cdots, k)$, which outputs new feature vectors with a size of $1 \times 1 \times 2c'$. These new feature vectors are followed by a sub-network consisting of a $1\times 1$ convolution, batch normalization, and ReLU layers. The function $\mathcal{F}(\hat{x}_i, \hat{x}_0, \{W_i\})$ is utilized to represent the concatenation operation and the following sub-network. The output of each $\mathcal{F}$ is a residual feature $r_i$ with a size of $1 \times 1 \times c'$. A shortcut connection \cite{RN73} is used to combine the residual feature $r_i$ and global feature $\hat{x}_0$ to generate the global-aware feature $x_i^g$,
	\begin{equation}
		x_i^g = \mathcal{F}(\hat{x}_i, \hat{x}_0, \{W_i\}) + \hat{x}_0, ~~~~i = (1,\cdots, k).
	\end{equation}

	The operation $\mathcal{F} + \hat{x}_0$ is achieved by an element-wise addition. 
	The size of global-aware feature $x_i^g$ is $1 \times 1 \times c'$.
	The weights of $k$ global-aware modules are not shared to ensure independent training for each aware module.

	The principle of the global-aware module is explained below. When dealing with the image with many background clutter and occlusions, the global-aware module explores the relationship between the global-level and local-level features using the function $\mathcal{F}(\hat{x}_i, \hat{x}_0, \{W_i\})$. Then a shortcut connection between the residual feature $r_i$ and global feature $\hat{x}_0$ is used to make the global-aware feature $x_i^g$ overcome the disadvantages of part models. 
	
	\subsection{Unsupervised Learning} \label{unsupervised learning}
	In this paper, a specific unsupervised method is not the emphasis. Therefore, the existing unsupervised method, \textit{i.e.,} Deep Association Learning (DAL) \cite{RN87}, is employed and incorporated into proposed scheme. Here, the formulation of this unsupervised module is presented using new notations.
	
	\paragraph{Anchor Update}
	An intra-camera anchor $\mathcal{I}$ is introduced to represent the feature center of each tracklet $T$, and the part intra-camera anchor $\mathcal{I}_i$ is employed to represent the part tracklet $T_i$. The part intra-camera anchor $\mathcal{I}_i$ is incrementally updated using Exponential Moving Average (EMA) strategy as follows,
	\begin{equation}
		\mathcal{I}^{t+1}_i = \mathcal{I}^{t}_i - \eta (\mathcal{I}^{t}_i - x_i^\Theta), ~~i = (1,\cdots, k), \label{intra-anchor update}
	\end{equation}
	where $i$ represent the $i^{th}$ part, $\eta$ refers to the update rate, and $t$ is the mini-batch learning iteration. $x^\Theta_i$ represents the local/global-aware feature. Since the camera and tracklet IDs are known, the features of the same body part from the person image within one tracklet $\{x_i^\Theta\}_{i=1}^k$ are used to form the part inter-camera anchor $\mathcal{I}_i$.
	
	The global Cyclic Ranking Consistency (CRC) \cite{RN187} is employed to explore the relationship between images from different cameras. For the specific $\mathcal{I}_i$ and $\tilde{\mathcal{I}_i}$, they are the same part of different tracklets and are from different camera views. The CRC means that the Euclidean distance between them is the smallest. Therefore, $\mathcal{I}_i$ and $\tilde{\mathcal{I}_i}$ is likely to be the same person from different cameras. Since $\mathcal{I}_i$ and $\tilde{\mathcal{I}_i}$ are highly related, a set of part cross-camera anchors $\mathcal{C}_i$ is introduced to denote their average feature representation, 
	\begin{equation}
		\mathcal{C}^{t+1}_i =   \left\{
		\begin{aligned}
			&\frac{1}{2}(\mathcal{I}^{t+1}_i + \tilde{\mathcal{I}}^{t}_i), && \mathrm{s.t.}~ {CRC}   \\
			&\mathcal{I}^{t+1}_i, && \mathrm{others,}
		\end{aligned}
		\right.                    \label{update cross-anchor}
	\end{equation}
	where $\mathrm{s.t.}$ is the abbreviation of \textit{subject to}.
	
	\paragraph{Distance Metric}
	In the training process, each mini-batch contains $M$ person images from different camera views. Each feature $x^\Theta_i$ represents a local/global-aware feature of one person image. This image is from one particular tracklet, \textit{i.e.,} its source tracklet. Therefore, each feature $x^\Theta_i$ has one source part intra-camera anchor $\mathcal{I}_i$. The Euclidean distances between each feature $x^\Theta_i$ and all part intra-camera anchors $\mathcal{I}_i$ from the same camera view are computed. The smallest distance of these Euclidean distances is denoted as $\mathrm{D}^{min}_i$, which means feature $x^\Theta_i$ belongs to this tracklet with a high possibility. Since there are many part intra-camera anchors $\mathcal{I}_i$, the distance between feature $x^\Theta_i$ and its source part intra-camera anchor $\mathcal{I}_i$ is denoted as $\mathrm{D}^\mathcal{I}_i$, which may not be the smallest. The distance $\mathrm{D}^\mathcal{C}_i$ can be obtained via the same operation by feature $x^\Theta_i$ and its source part cross-camera anchor $\mathcal{C}_i$. For the images from the same camera view in a mini-batch, the average value of the smallest distances $\mathrm{D}^{min}_i$ is denoted as $\bar{\mathrm{D}_i}$. This average value $\bar{\mathrm{D}_i}$ is used to ensure each feature has the same distance with its feature center, \textit{i.e.,} its source part intra-camera anchor $\mathcal{I}_i$. The $k$ unsupervised modules are trained individually. Therefore, the distances above are calculated using the $i^{th}$ feature and its corresponding part intra/cross-camera anchors.
	
	\paragraph{Association Loss}
	Two top-push margin-based association losses are used. The first one is intra-camera association loss,
	\begin{equation}
		L^\mathcal{I}_i = \left\{
		\begin{aligned}
			&[\mathrm{D}^\mathcal{I}_i - \mathrm{D}^{min}_i + m]_+, && \mathrm{D}^\mathcal{I}_i \neq \mathrm{D}^{min}_i  \\
			&[\mathrm{D}^\mathcal{I}_i - \bar{\mathrm{D}}_i + m]_+, && \mathrm{D}^\mathcal{I}_i = \mathrm{D}^{min}_i,
		\end{aligned}
		\right. \label{loss1}
	\end{equation}
	where $[\bullet]_+ = \text{max}(0, \bullet)$, and $m$ is the margin enforcing the deep model to assign the source part tracklet as the top-rank. The second one is the cross-camera association loss,
	\begin{equation}
		L^\mathcal{C}_i = \left\{
		\begin{aligned}
			&[\mathrm{D}^\mathcal{C}_i - \mathrm{D}^{min}_i + m]_+, && \mathrm{D}^\mathcal{I}_i \neq \mathrm{D}^{min}_i  \\
			&[\mathrm{D}^\mathcal{C}_i - \bar{\mathrm{D}_i} + m]_+, && \mathrm{D}^\mathcal{I}_i = \mathrm{D}^{min}_i.
		\end{aligned}
		\right. \label{loss2}
	\end{equation}
	The goal of $L^\mathcal{I}_i$ is to associate feature $x^\Theta_i$ with its source part intra-camera anchor $\mathcal{I}_i$ at a proper distance. Meanwhile, $L^\mathcal{C}_i$ is to pull the part intra-camera anchor $\mathcal{I}_i$ close to its CRC part intra-camera anchor $\tilde{\mathcal{I}_i}$. The final learning objective for this unsupervised module is to optimize two association losses jointly,
	\begin{equation}
		L^u_i = L^\mathcal{I}_i + \lambda L^\mathcal{C}_i,
	\end{equation}
	where $\lambda$ is a trade-off parameter balancing these two losses. Besides, other unsupervised methods can be employed to deal with the different features $x^\Theta_i$ in the same manner, which makes proposed scheme a general one. 
	
	Then $k$ learning objectives $L_i^u$ are employed to calculate the final loss $L$,
	\begin{equation}
		L = \frac{1}{k}\sum_{i=1}^k L_i^u. \label{average intra anchor}
	\end{equation}
		
	\subsection{Model Training and Testing} \label{model training and testing}
	In the training process, there are $n$ warmup epochs, where the part cross-camera anchor $\mathcal{C}_i$ is always set as its corresponding $\mathcal{I}_i$ as following,
	\begin{equation}
		\mathcal{C}^{t}_i = \mathcal{I}^{t}_i, ~~~t~\mathrm{in}~n~\mathrm{epoches}. 
	\end{equation}
	After $n$ warmup epochs, the part cross-camera anchors are updated using Eq. \ref{update cross-anchor} for the rest of the epochs. This trick ensures that the model learns intra-camera association in warmup epochs and then learns intra-camera and cross-camera association jointly. 
	
	The proposed scheme is trained twice where local-aware module and global-aware module are incorporated, respectively. In each training stage, proposed scheme learns different perceptive abilities. The scheme using local-aware module receives a smaller receptive field for each part and has an unstable performance. The scheme using global-aware module explores the relationship between the local-level feature and global-level feature, which provides a general performance improvement. The local-aware module is based on local-level features, and the global-aware module is based on global-level features. 
	
\begin{figure}[!h]
		\centering{\includegraphics[width=0.48\textwidth]{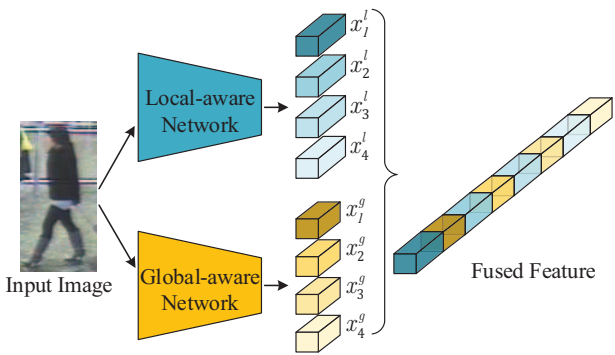}}
		\caption{Illustration of the fusion process. After the training process, the proposed scheme incorporating the local-aware module is denoted as the local-aware network. The global-aware network incorporates the global-aware module. The partition scale $k$ is set to 4 in this diagram.  \label{framework-for-test}} 
	\end{figure}

	After the training process, the features from these two modules are fused to form a robust feature representation for each input image, as illustrated in Fig. \ref{framework-for-test}. Concretely, each image is represented by $k$ local-aware features with a size of $1\times 1\times c$ and $k$ global-aware features with a size of $1\times 1\times c'$. Then the corresponding ones, $x_i^l$ and $x_i^g$, are concatenated to form $k$ features $\{x_i\}_{i=1}^k$ with a size of $1\times 1\times (c+c')$. After normalization, $k$ features $\{x_i\}_{i=1}^k$ are concatenated to form the robust feature representation with a size of $1\times 1\times k(c+c')$ for each image. Each tracklet consists of $N$ person images. Thus the size of the feature for each tracklet is $N\times k(c+c')$. A \texttt{max} operation is applied to this feature to select the most discriminative value along the $N$ images to generate the feature representation with a size of $1\times k(c+c')$. After normalization, this generated feature is regarded as the feature of each tracklet for the test.
	
	\section{Experiments} \label{Experiments}
	Comprehensive experiments are conducted to validate the effectiveness of proposed uPMnet. Experimental settings, experimental results, ablation study are given in the following parts.
	
	\subsection{Experimental Settings}
	\paragraph{Datasets}
	The proposed uPMnet is evaluated on three video-based reID benchmarks.
	\begin{itemize}
		\item PRID2011 \cite{RN386} is captured by two non-overlapping surveillance cameras on campus. It includes 385 tracklets in camera A and 749 tracklets in camera B. In this dataset, 200 persons appear in both camera views, which outputs 400 tracklets with an average length of 100 frames. The experimental setting of \cite{RN386} \cite{RN87} \cite{RN109} \cite{RN102} is followed, and only 178 identities are used, where each tracklet has more than 27 frames.
		\item iLIDS-VID \cite{RN385} has been collected from two disjoint cameras at an airport hall. It consists of 600 tracklets of 300 persons. There are two sequences from different camera views for each person, where each tracklet has an average duration of 73 frames. It is a quite challenging benchmark due to significant lighting, viewpoint variations, severe background clutter, and mutual occlusions.
		\item DukeMTMC-VideoReID is derived from the image-based benchmark, DukeMTMC \cite{RN170}, and re-organized by Wu \textit{et al.} \cite{RN391}. DukeMTMC-VideoReID contains 4832 tracklets and 1812 identities in total, and each tracklet has 168 frames on average.
	\end{itemize}

	\begin{table*}[!ht]
		\centering
		\begin{threeparttable}
			\caption{Performance comparisons with other state-of-the-art unsupervised video reID methods on PRID2011, iLIDS-VID and DukeMTMC-VideoReID.}\label{SOTA}
			\centering
			\begin{tabular}{lccccccccccccc}
				\hline
				\multirow{2}{*}{Methods}       & \multirow{2}{*}{Venue} & \multirow{2}{*}{Backbone}  & \multicolumn{3}{c}{PRID2011} & \multicolumn{3}{c}{iLIDS-VID}  & \multicolumn{4}{c}{DukeMTMC-VideoReID}\\
				\cmidrule(r){4-6}  \cmidrule(r){7-9} \cmidrule(r){10-13} 
				&            &            & R1       & R5  & R20  & R1   & R5   & R20  & R1 & R5 & R20 & mAP    \\ \hline
				OIM* \cite{RN248}                                                   & CVPR2017                   & ResNet50                   & -                      & -                      & -                      & -                               & -                      & -                      & 51.1                   & 70.5                   & 76.2                & 43.8                    \\
				TAUDL \cite{RN369}                                & ECCV2018  & ResNet50 & 49.4 & 78.7 & 98.9 & 26.7          & 51.3 & 82.0 & -    & -    & -    & -      \\
				DAL \cite{RN87}                                                     & BMVC2018                   & MobileNet                  & 84.6                   & 96.3                   & 99.1                   & 52.8                            & 76.7                   & 91.6                   & -                      & -                      & -                   & -                       \\
				DAL \cite{RN87}                                                     & BMVC2018                   & ResNet50                   & 85.3                   & 97.0                   & 99.6                   & 56.9                            & 80.6                   & 91.9                   & -                      & -                      & -                   & -                       \\
				BUC \cite{RN247}                                                    & AAAI2019                   & ResNet50                   & -                      & -                      & -                      & -                               & -                      & -                      & 69.2                   & 81.1                   & 85.8                & 61.9                    \\
				UGA \cite{RN370}                                                    & ICCV2019                   & ResNet50                   & 80.9                   & 94.4                   & 100                    & 57.3                            & 72.0                   & 87.3                   & -                      & -                      & -                   & -                       \\
				DBC \cite{RN118}                                  & BMVC2019  & ResNet50 & -    & -    & -    & -             & -    & -    & 75.2 & 87.0 & - & 66.1  \\ 
				
				UTAL \cite{RN97}                                                    & TPAMI2019                  & ResNet50                   & 54.7                   & 83.1                   & 96.2                   & 35.1                            & 59.0                   & 83.8                   & -                      & -                      & -                   & -                       \\
				TSSL \cite{RN389}                                                   & AAAI2020                   & ResNet50                   & -                      & -                      & -                      & -                               & -                      & -                      & 73.9                   & -                      & -                   & 64.6                    \\
				SSL \cite{RN492}                                                    & CVPR2020                   & ResNet50                   & -                      & -                      & -                      & -                               & -                      & -                      & 76.4                   & 88.7                   & -                   & 69.3                    \\   
				NHAC \cite{RN536}                                 & arXiv2021 & ResNet50 & -    & -    & -    & -             & -    & -    & 82.8 & 92.7 & - & 76.0  \\ \hline
				\multicolumn{2}{l}{Proposed uPMnet}                                                              & ResNet50                   & \textbf{92.0}          & 97.7                   & \textbf{100}           & \textbf{63.1}                   & \textbf{81.9}          & \textbf{92.5}          & 81.3                   & 91.7                   & 97.2                & 74.6                    \\
				\multicolumn{2}{l}{Proposed uPMnet}                                                              & MobileNet                  & 90.2                   & \textbf{97.8}          & 100                    & 62.6                            & 80.9                   & 91.6                   & \textbf{83.6}          & \textbf{93.1}          & \textbf{97.2}       & \textbf{76.9}           \\ \hline 
				AGRL\dag \cite{RN442} (supervised)                                  & TIP2020                    &                            & 94.6                   & 99.1                   & 100                    & 84.5                            & 96.7                   & 99.5                   & 97.0                   & 99.3                   & 99.9                & 95.4                    \\ \hline
			\end{tabular}
			\begin{tablenotes}
				\item[1] R1, R5 and R20 are the abbreviations of Rank-1, Rank-5 and Rank-20, respectively.
				\item[2] `-': no reported results. `*': reported in \cite{RN247}.
				\item[3] The best results are in \textbf{bold}. 
				\item[4] `\dag' : supervised method, AGRL, is listed as an upper bound of performance on each benchmark.
				\item[5] On each benchmark, the reported best results of each method are listed.
			\end{tablenotes}
		\end{threeparttable}
	\end{table*}
	
	\paragraph{Evaluation Protocol}
	For PRID2011 and iLIDS-VID benchmarks, the whole set of tracklet pairs are randomly divided into two halves for training and testing in multiple trials. These tracklets of the same person from two camera views compose the probe set and gallery set, respectively. The trials are repeated ten times to ensure a statistically stable result. For DukeMTMC-VideoReID benchmarks, the training and testing split manners are followed the experimental setting of \cite{RN391} and \cite{RN158}. 
	
	\paragraph{Metrics}
	The performances are measured by average Cumulated Matching Characteristics (CMC) curves. The Rank-1, Rank-5, Rank-20 scores are employed to represent the CMC curve. For DukeMTMC-VideoReID benchmarks, the mean Average Precision (mAP) are also used to measure the performance. For PRID2011 and iLIDS-VID benchmarks, each person ID has only one correct tracklet in the gallery. In other words, each person ID has only one precision value. It is not necessary to calculate the "average precision (AP)". And the CMC metric is enough to measure the ranked gallery list. Therefore, the mAP metric is not used for these two benchmarks.
	
	\paragraph{Implementation Details}
	The proposed uPMnet is implemented using Tensorflow \cite{RN398}. For PRID2011 and iLID-VID, RMSProp with an initial learning rate of $4.5\times10^{-2}$ is used to train the models $2\times10^4$ and $1.5\times10^4$ iterations, respectively. For DukeMTMC-VideoReID, standard stochastic gradient descent (SGD) with learning rate initialized to $1\times10^{-2}$ with momentum is used to train the models $2.5\times10^4$ iterations. For all datasets, the batch size $M$ is set to 64, and person images are resized to $256 \times 128$. 
	The pre-trained ResNet50 \cite{RN73} and  MobileNet \cite{RN228} are employed as the backbone. The dimensions (height $\times$ width $\times$ channel) of feature map outputted from ResNet50 and MobileNet are $8\times 4\times 2048$ and $8\times 4 \times 1024$, respectively. 
	The update rate $\eta$ and the margin $m$ are empirically set to 0.5, and the trade-off parameter $\lambda$ is set to 1. The reduced feature channel $c'$ for the global-aware module is set to 256. The warmup epoch $n$ is set to 2 and 1 for small-scale datasets (PRID2011 and iLIDS-VID) and large-scale dataset (DukeMTMC-VideoReID), respectively. The average feature representation of all person images within the same tracklet is used as the tracklet feature. Although previous existing literature \cite{RN92} \cite{RN533} proposed to assign different weights for each person image from the same tracklet, studying the adaptive weight is not the emphasis of this paper. 
	
	\subsection{Comparison with State-of-the-art}
	Table \ref{SOTA} illustrates the comparison results between proposed uPMnet with other state-of-the-art methods. These methods are OIM \cite{RN248}, TAUDL \cite{RN369}, DAL \cite{RN87}, BUC \cite{RN247}, UGA \cite{RN370}, DBC \cite{RN118}, UTAL \cite{RN97}, TSSL \cite{RN389}, SSL \cite{RN492}, NHAC \cite{RN536}. A supervised method, AGRL \cite{RN442}, is also listed as an upper bound of performance on each benchmark. Among all these existing methods, proposed uPMnet is the only one using part models-based CNN model. On each benchmark, the reported best results of each method are listed. As illustrated in Table \ref{SOTA}, proposed uPMnet with different backbones achieve the first and second place alternatively on these three benchmarks.
	
	\begin{figure}
		\centering
		\includegraphics[width=0.475\textwidth]{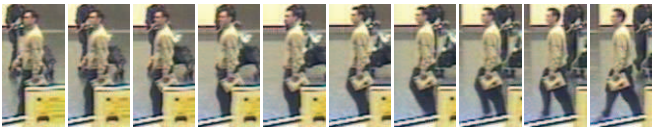}
		\caption{Background clutter and mutual occlusions are pervasive in iLIDS-VIS benchmark.} \label{iLIDS}
	\end{figure}
	
	\paragraph{Results on RPID2011}
	Compared to the other state-of-the-art approaches on PRID2011, proposed uPMnet achieves the best performance. There is a large Rank-1 improvement of 6.7\% (92.0-85.3) between proposed uPMnet and the best competitor, DAL (ResNet50) \cite{RN87}. In this paper, DAL is the baseline method, which uses two margin-based association losses to constrain each person image feature to its source intra-camera and cross-camera feature center. The DAL uses the global-level feature only. The significant performance improvement demonstrates the effectiveness of proposed uPMnet. Besides, proposed uPMnet is the first model that achieves above 90.0\% on Rank-1, which is very close to the result of supervised method AGRL \cite{RN442}.
		
	\paragraph{Results on iLIDS-VID}
	Table \ref{SOTA} illustrates that proposed uPMnet achieves the best performance on iLIDS-VID benchmark. Compared to PRID and DukeMTMC-VideoReID benchmarks, iLIDS-VID is a challenging one due to its surveillance environment. As illustrated in Fig. \ref{iLIDS}, background clutter and mutual occlusions are pervasive in iLIDS-VID benchmark \cite{RN385}. However, compared to the other state-of-the-art approaches, proposed uPMnet still achieves improvement by a large Rank-1 improvement of 5.8\% (63.1-57.3) over the best competitor, UGA \cite{RN370}. The UGA employed a two-stage training manner to learn a view-invariant feature representation from the possible images. While proposed uPMnet introduces a robust part models-based scheme and train the model in an end-to-end fashion. This comparison demonstrates that part models with elaborate design can achieve a robust performance for the unsupervised video reID task.

	\paragraph{Results on DukeMTMC-VideoReID}
	On DukeMTMC-VideoReID, proposed uPMnet also achieves state-of-the-art performance. Compared to PRID2011 and iLIDS-VID, DukeMTMC-VideoReID is a large-scale benchmark with 4832 tracklets in total, which is eight times of iLIDS-VID (600 tracklets) and thirteen times of PRID2011 (356 tracklets). The proposed uPMnet achieves performance improvement by a Rank-1 improvement of 0.8\% (83.6-82.8) and an mAP improvement of 0.9\% (76.9-76.0) over the best competitor, NHAC \cite{RN536}. The NHAC method proposed a graph trimming module and a node re-sampling module to improve the clustering-based unsupervised learning method.
	Proposed uPMnet depends on the camera ID to explore the intra-camera and cross-camera association learning and achieves better performance in this large-scale benchmark.
	
	\begin{table*}[!h]
		\centering
		\begin{threeparttable}
			\caption{Performance comparisons for proposed uPMnet with different backbones on PRID2011, iLIDS-VID, and DukeMTMC-VideoReID.}\label{backbones}
			\begin{tabular}{ccccccccccccc}
				\hline
				\multirow{2}{*}{Backbone} & \multirow{2}{*}{Local-aware} & \multirow{2}{*}{Global-aware} & \multicolumn{3}{c}{PRID2011} & \multicolumn{3}{c}{iLIDS-VID} & \multicolumn{4}{c}{DukeMTMC-VideoReID} \\
				\cmidrule(r){4-6}  \cmidrule(r){7-9} \cmidrule(r){10-13}
				&            &              & Rank-1 & Rank-5 & Rank-20 & Rank-1 & Rank-5 & Rank-20 & Rank-1 & Rank-5 & Rank-20 & mAP  \\ \hline
				\multirow{4}{*}{MobileNet} &            &      & 82.3   & 96.9   & 99.6    & 54.1   & 76.9   & 90.7    & 79.3   & 92.5   & 96.7    & 74.2 \\
				& \checkmark   &  & \textbf{91.5}   & \textbf{98.3}   & 99.7   & 52.4   & 74.1   & 88.4    & 81.4   & 92.7   & 97.0    & 75.5 \\
				&  & \checkmark & 83.3   & 96.0   & 99.6    & 58.2   & 80.0   & 91.3    & 80.3   & 92.3   & 96.0    & 73.9 \\
				& \checkmark  & \checkmark & 90.2   & 97.8   & \textbf{100}    & \textbf{62.6}   & \textbf{80.9}   & \textbf{91.6}    & \textbf{83.6}   & \textbf{93.1}   & \textbf{97.2}    & \textbf{76.9} \\ \hline
				\multirow{4}{*}{ResNet50}  &    &              & 83.9   & 95.9   & 99.7    & 54.8   & 76.6   & 89.9    & 74.2   & 88.1   & 94.8    & 66.0 \\
				& \checkmark   & & 90.4   & 97.5   & 99.7    & 55.5   & 78.3   & 92.0    & 78.4   & 90.1   & 96.0    & 70.3 \\
				&  & \checkmark & 84.3   & 95.9   & 99.6    & 60.1   & 81.1   & 91.6    & 76.7   & 88.6   & 94.5    & 68.5 \\ 
				& \checkmark & \checkmark & \textbf{92.0}   & \textbf{97.7}   & \textbf{100}    & \textbf{63.1}   & \textbf{81.9}   & \textbf{92.5}    &  \textbf{81.3}  & \textbf{91.7}   & \textbf{97.2}    & \textbf{74.6} \\\hline
			\end{tabular}
			\begin{tablenotes}
				\item[1] The partition scale $k$ is set to 8.
				\item[2] For different backbone, the best results are in \textbf{bold}.
				\item[3] The performance of proposed uPMnet are listed at the last row for each backbone.
			\end{tablenotes}
		\end{threeparttable}
	\end{table*}

	\subsection{Ablation Study} \label{sec-ablation-study}
	The ablation study for proposed scheme with different backbones is illustrated in Table \ref{backbones}. The baseline method is a standard DAL \cite{RN187} and is a special case of proposed scheme. For the baseline method, only the global-level feature $x_0$ is used, and there is only one unsupervised module to operate this feature $x_0$.
	
	\begin{table*}[!h]
		\centering
		\begin{threeparttable}
			\caption{Performance comparisons of uPMnet with GAP and GMP on PRID2011, iLIDS-VID, and DukeMTMC-VideoReID.}\label{GAP}
			\begin{tabular}{cccccccccccc}
				\hline
				\multicolumn{2}{c}{\multirow{2}{*}{\begin{tabular}[c]{@{}c@{}}Proposed uPMnet\\ (MobileNet, $k$=2)\end{tabular}}} & \multicolumn{3}{c}{PRID2011} & \multicolumn{3}{c}{iLIDS-VID}  & \multicolumn{4}{c}{DukeMTMC-VideoReID} \\
				\cmidrule(r){3-5}  \cmidrule(r){6-8} \cmidrule(r){9-12} 
				\multicolumn{2}{c}{}              & Rank-1            & Rank-5   & Rank-20            & Rank-1   & Rank-5   & Rank-20  & Rank-1            & Rank-5            & Rank-20  & mAP       \\ \hline 
				& w/ GAP & 85.1  & \textbf{97.4} & \textbf{99.8}  & \textbf{58.7} & \textbf{78.5} & \textbf{90.8}  & \textbf{80.7}   & \textbf{92.7} & \textbf{96.8} & \textbf{74.9}    \\ 
				& w/ GMP & \textbf{85.8}    & 96.7  & 99.6 & 51.9 & 75.3 & 89.2  & 75.7 & 89.1 & 94.8          & 69.0    \\ \hline
			\end{tabular}
			\begin{tablenotes}
				\item[1] For rapid deployment, MobileNet is employed as backbone and $k$ is set to 2.
			\end{tablenotes}
		\end{threeparttable}
	\end{table*}

	\paragraph{Effectiveness of Proposed Local-aware Module}
	In this section, proposed scheme with the local-aware module is analyzed. When using MobileNet as the backbone, proposed scheme achieves an average Rank-1 improvement of 5.7\% and mAP improvement of 1.3\% on the PRID2011 and DukeMTMC-VideoReID benchmarks compared to the baseline. The performance gap can be up to 9.2\% (91.5-82.3, the Rank-1 gap on PRID2011 benchmark compared to the baseline). When using ResNet50 as the backbone, proposed scheme achieves an average Rank-1 improvement of 3.8\% and mAP improvement of 4.3\% on three benchmarks compared to the baseline method. Since the local-aware module only employs the local-level information, these performance improvements demonstrate that part models can improve the performance for unsupervised learning to some extent.
	
	For the iLIDS-VID benchmark, proposed scheme using MobileNet as backbone performs worse than the baseline method. Although proposed scheme using ResNet50 as backbone achieves performance improvement, the performance improvement is much less than the performance improvement on PRID and DukeMTMC-VideoReID, \textit{i.e.,} Rank-1 improvement of 0.7\% on iLIDS-VID vs. average Rank-1 improvement of 5.4\% on PRID and DukeMTMC-VideoReID. The main reason mainly comes from the nature of the iLIDS-VID benchmark, where background clutters and mutual occlusions are pervasive \cite{RN385} \cite{RN96}, as illustrated in Fig. \ref{iLIDS}. In this situation, the part models produce more similar feature representations than the global one, which makes the unsupervised method hard to distinguish the same body part of different persons. In this situation, the negative effects of part models are amplified remarkably.

	\paragraph{Effectiveness of Proposed Global-aware Module}
	In this section, proposed scheme with the global-aware module is analyzed. For all the benchmarks, proposed scheme using MobileNet as backbone achieves an average Rank-1 improvement of 2.0\% compared to the baseline method. Proposed scheme using ResNet50 as backbone achieves an average Rank-1 improvement of 2.7\%. Proposed global-aware module explores the relationship between the local-level features and global-level features and overcomes the disadvantages of part models. These comparisons demonstrate the effectiveness of proposed global-aware module.
	
	\paragraph{Effectiveness of Proposed Feature Fusion}
	The performance differences on three benchmarks are analyzed here. Either using MobileNet as backbone or using ResNet50 as backbone, proposed scheme using the global-aware module achieves more obvious improvement on the iLIDS-VID benchmark, \textit{i.e.,} average Rank-1 improvement of 4.7\% on iLIDS-VID vs. average Rank-1 improvement of 1.2\% on PRID and DukeMTMC-VideoReID. Meanwhile, proposed scheme using local-aware module achieves more obvious improvement on PRID and DukeMTMC-VideoReID than on iLIDS-VID. The local-aware module and global-aware module have complimentary performances. 
	
	Table \ref{backbones} illustrates the results of feature representation fused from the local-aware feature and the global-aware feature. The fused feature obtains first place for almost all metrics. Through the feature fusion, the complementary performances are combined decently, and the performance is improved further. The fused feature absorbs the superiority of part models and discards its shortcomings for various challenging situations. These comparisons demonstrate the effectiveness of the feature fusion operation. 
	
	\subsection{Global Average Pooling vs. Global Max Pooling} \label{GAP vs. GMP}
	Many supervised methods \cite{RN318} \cite{RN394} \cite{RN314} proved that global max pooling (GMP) is more effective than global average pooling (GAP). In the second part of proposed scheme, \textit{i.e., part/global-aware feature generation}, GAP, not GMP, is used to generate the \textit{compact feature}. The performance comparisons are listed in Table \ref{GAP}. Compared to the scheme with GAP, the scheme with GMP is almost inferior on each metric. These comparisons are explained below. Supervised methods supervise the model to learn discriminative features to distinguish different persons, while GMP can aggregate the feature from the most discriminative part. Unsupervised methods focus on the undiscriminating feature because they mostly rely on the similarities to distinguish possible pair features, while GAP introduces an undiscriminating feature representation covering the whole person image. Thus employing GAP is a better choice for unsupervised learning.
	
	\begin{table*}[!h]
		\centering
		\begin{threeparttable}
			\caption{Performance comparisons of uPMnet with different partition scales on PRID2011, iLIDS-VID, and DukeMTMC-VideoReID.}\label{partK}
			\begin{tabular}{cccccccccccc}
				\hline
				\multicolumn{2}{c}{\multirow{2}{*}{MobileNet}} & \multicolumn{3}{c}{PRID2011} & \multicolumn{3}{c}{iLIDS-VID}  & \multicolumn{4}{c}{DukeMTMC-VideoReID} \\
				\cmidrule(r){3-5}  \cmidrule(r){6-8} \cmidrule(r){9-12} 
				\multicolumn{2}{c}{}              & Rank-1            & Rank-5   & Rank-20            & Rank-1   & Rank-5   & Rank-20  & Rank-1            & Rank-5            & Rank-20  & mAP       \\ \hline
				{Baseline}              & $k=1$ & 82.3          & 96.9 & 99.6           & 54.1 & 76.9 & 90.7  & 79.3          & 92.5          & 96.7          & 74.2    \\ \hline
				\multirow{3}{*}{\begin{tabular}[c]{@{}c@{}}Proposed uPMnet\\ \end{tabular}}  & $k=2$ & 85.1          & 97.4 & 99.8           & 58.7 & 78.5 & 90.8  & 80.7          & 92.7          & 96.8          & 74.9    \\
				& $k=4$ & 88.8          & 97.8  & 100 & 58.6 & 79.6 & 91.1  & 82.6 & \textbf{93.3} & \textbf{97.4}          & 75.6    \\
				& $k=8$ & \textbf{90.2} & \textbf{97.8} & \textbf{100.0} & \textbf{62.6} & \textbf{80.9} & \textbf{91.6}  & \textbf{83.6} & 93.1 & 97.2 & \textbf{76.9} \\ \hline
			\end{tabular}
			\begin{tablenotes}
				\item[1] MobileNet is employed as backbone for rapid deployment.
			\end{tablenotes}
		\end{threeparttable}
	\end{table*}

	\subsection{Parameter Analysis}
	Table \ref{partK} shows the performances of proposed uPMnet using different partition scales. The MobileNet is selected as the backbone for rapid deployment. 
	Since the height of feature map is 8, the partition scale $k$ is set to 1,2,4,8 to take the experiments. 
	With the increase of partition scale $k$, \textit{e.g.}, from 1 to 8, proposed uPMnet improves apparently. The average gain between adjacent scales, \textit{e.g.}, 2 vs. 1, 4 vs. 2 and 8 vs. 4, is 2.2\% for Rank-1 on these three benchmarks. These comparisons demonstrate that a more fine-grained feature can optimize the performance of the unsupervised method. Since the Rank-1 metric represents the model ability to recognize the easiest person images, proposed uPMnet can effectively improve this ability. For the metric of mAP, the average gain between adjacent scales is 0.9\% on the DukeMTMC-VideoReID benchmark. These comparisons demonstrate the effectiveness of proposed scheme.
	
	\begin{figure}[]
		\centering
		\begin{minipage}[b]{0.475\textwidth} 
			\includegraphics[width=1\textwidth]{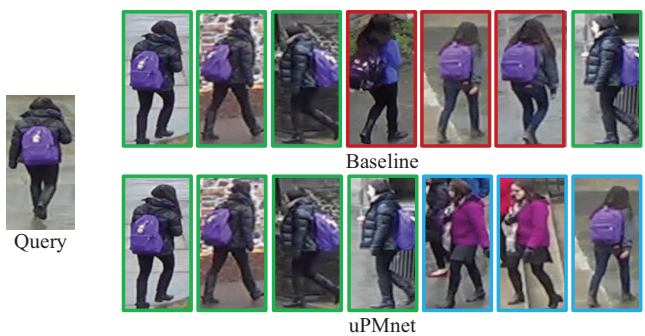}
		\end{minipage}
		\caption{The top-7 retrieval tracklets of baseline and proposed uPMnet are shown here. Each tracklet is represented by one image randomly sampled from this tracklet for convenience. 
		} \label{qualitative results}
	\end{figure}
	
	\begin{figure}[]
		\centering
		\subfigure[Examples of activation maps on PRID2011.]{
			\begin{minipage}{0.21\textwidth} \label{cam PRID}
				\centering
				\includegraphics[scale=0.18]{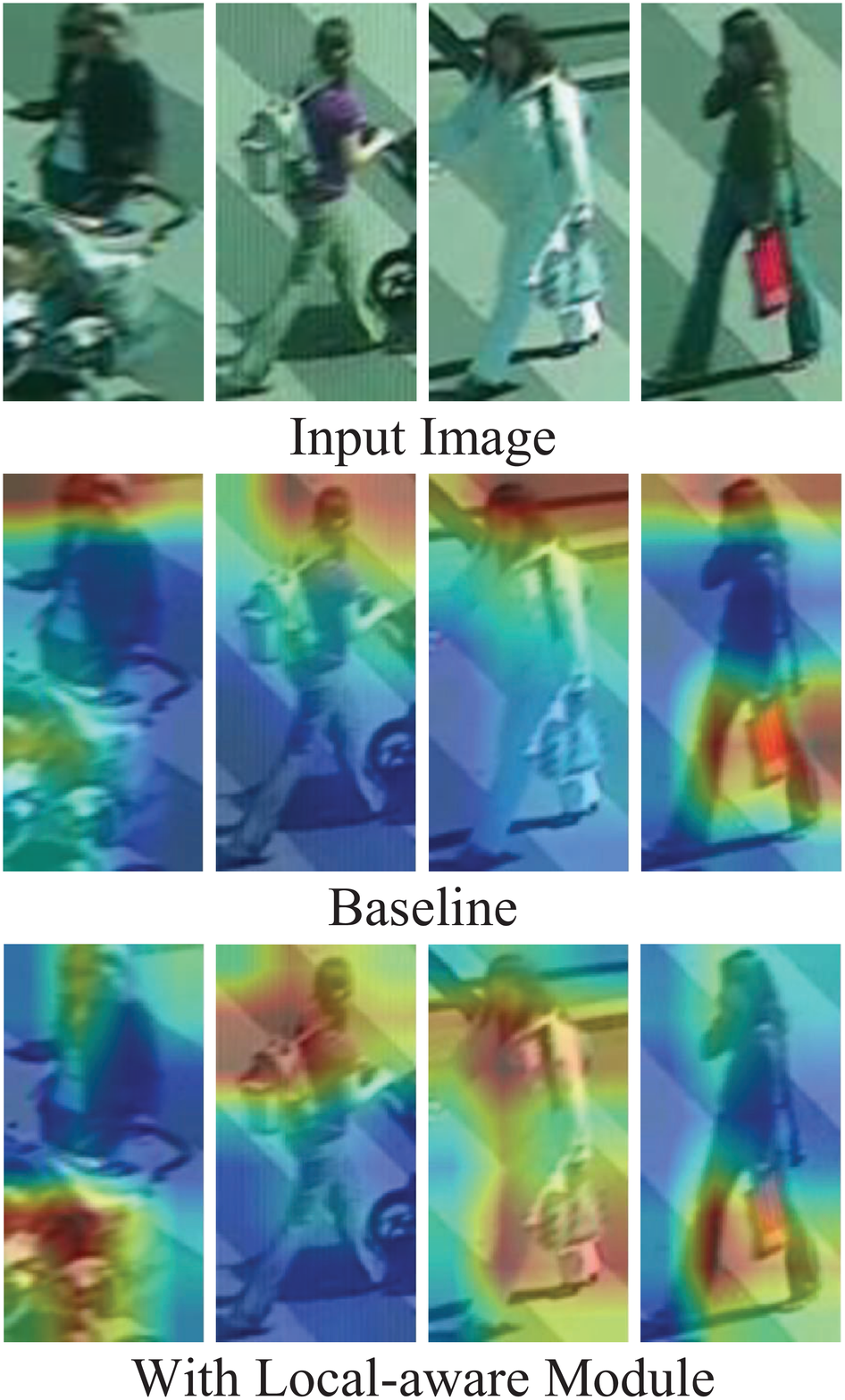}
		\end{minipage}}\hspace{+5mm}
		\subfigure[Examples of activation maps on DukeMTMC-VideoReID.]{
			\begin{minipage}{0.21\textwidth} \label{cam Duke}
				\centering
				\includegraphics[scale=0.18]{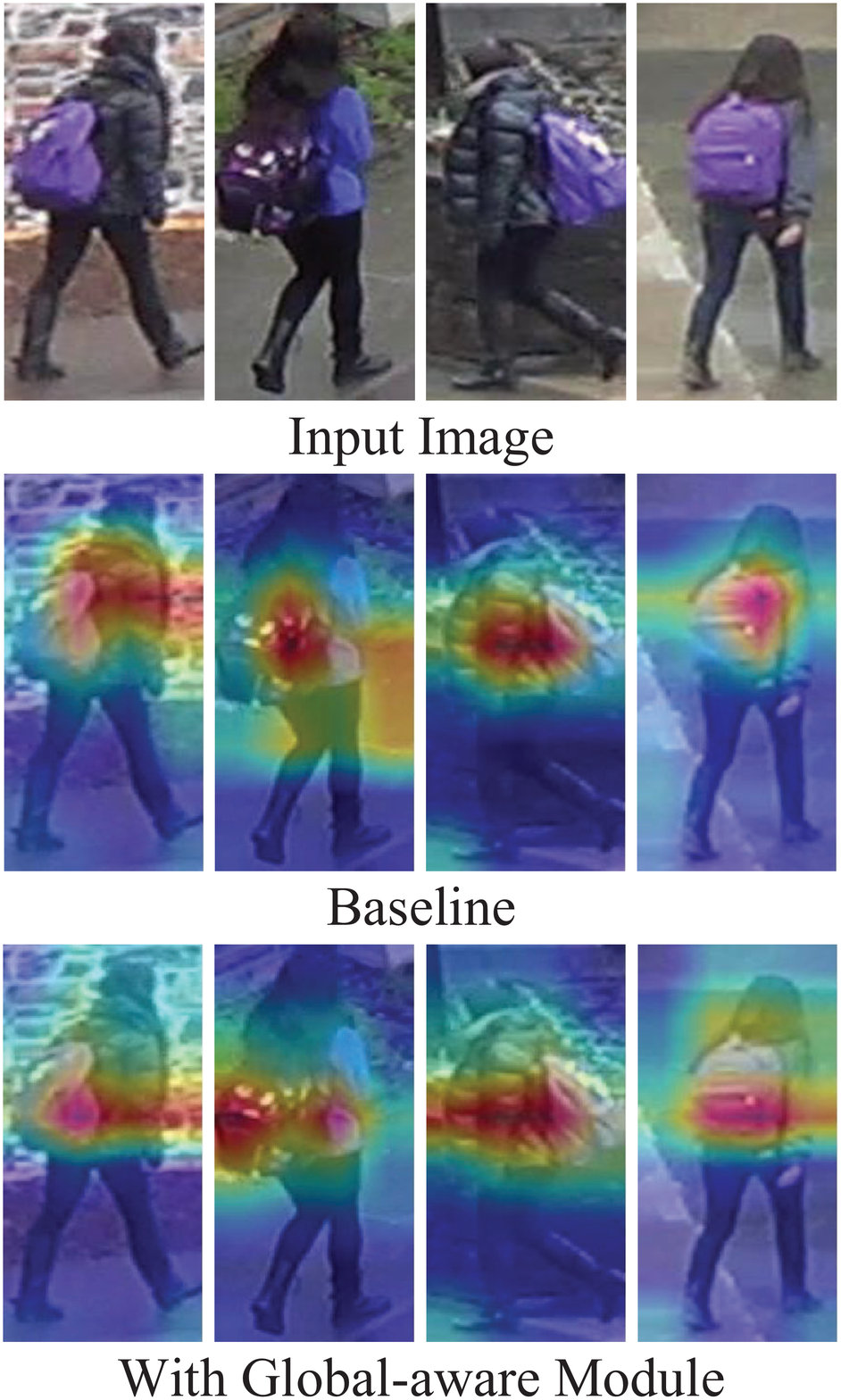}
		\end{minipage}}
		\caption{Activation maps of the baseline and proposed scheme using different aware modules. The first row is the randomly sampled images. The class activation maps on PRID2011 and DukeMTMC-VideoReID benchmarks are illustrated.} \label{cam}
	\end{figure}

	\begin{table*}[h]
		\centering
		\begin{threeparttable}
			\caption{Computational Complexity Comparisons on PRID2011, iLIDS-VID, and DukeMTMC-VideoReID.}\label{Computational3}
			\scriptsize
			\begin{tabular}{clllllllll}
				\hline
				
				\multirow{2}{*}{Method} & \multicolumn{3}{c}{PRID2011} & \multicolumn{3}{c}{iLIDS-VID}  & \multicolumn{3}{c}{DukeMTMC-VideoReID} \\
				\cmidrule(r){2-4} \cmidrule(r){5-7} \cmidrule(r){8-10}
				                                   & FLOPs                          & Params                          & Rank-1                       & FLOPs                          & Params                          & Rank-1                        & FLOPs                           & Params                          & Rank-1                       \\ \hline
				Baseline                           & 4.6G                           & 23.8M                           & 83.9                         & 4.6G                           & 24.1M                           & 54.8                          & 4.7G                            & 28.0M                           & 74.2                        \\
				Local-aware                        & 4.7G\tiny{($\uparrow$0.8\%)}   & 26.4M\tiny{($\uparrow$10.7\%)}  & 90.4\tiny{($\uparrow$7.7\%)} & 4.7G\tiny{($\uparrow$1.4\%)}   & 28.4M\tiny{($\uparrow$17.9\%)}  & 55.5\tiny{($\uparrow$1.2\%)}  & 5.2G\tiny{($\uparrow$10.6\%)}   & 59.5M\tiny{($\uparrow$112.5\%)} & 78.4\tiny{($\uparrow$5.6\%)}\\
				Global-aware                       & 4.7G\tiny{($\uparrow$0.8\%)}   & 29.6M\tiny{($\uparrow$24.1\%)}  & 84.3\tiny{($\uparrow$0.4\%)} & 4.7G\tiny{($\uparrow$0.8\%)}   & 29.9M\tiny{($\uparrow$23.9\%)}  & 60.1\tiny{($\uparrow$9.6\%)}  & 4.7G\tiny{($\uparrow$0.8\%)}    & 33.7M\tiny{($\uparrow$20.6\%)}  & 76.7\tiny{($\uparrow$3.3\%)}\\
				Proposed uPMnet                    & 9.4G\tiny{($\uparrow$101.7\%)} & 56.0M\tiny{($\uparrow$134.9\%)} & 92.0\tiny{($\uparrow$9.6\%)} & 9.4G\tiny{($\uparrow$102.3\%)} & 58.3M\tiny{($\uparrow$141.8\%)} & 63.1\tiny{($\uparrow$15.1\%)} & 10.0G\tiny{($\uparrow$111.4\%)} & 93.3M\tiny{($\uparrow$233.1\%)} & 81.3\tiny{($\uparrow$9.5\%)}\\ \hline
			\end{tabular}
			\begin{tablenotes}
				\item[1] `$\uparrow$': Means the percentage increase relative to the baseline method.
				\item[2] ResNet50 is employed as backbone in this table.
			\end{tablenotes}
		\end{threeparttable}
	\end{table*}

	\subsection{Qualitative Results}
	Fig. \ref{qualitative results} shows visual comparisons of person retrieval results on large-scale benchmarks, DukeMTMC-VideoReID. After spotting all correct ones, the incorrect ones are deliberately kept as negative ones. The images with green, red, and blue boxes are correct, incorrect, and negative ones. The difference between the baseline and proposed uPMnet can be observed. For the red ones at the first row in Fig. \ref{qualitative results}, the baseline approach spots the assigned person by similar contour, which results in obvious mistakes. Meanwhile, proposed method recognizes all the correct person images. These comparisons demonstrate that proposed uPMnet can achieve more discriminative features than the baseline method.
	
	Fig. \ref{cam} visualizes the activation maps and illustrate the different effects of using local-aware module and global-aware module. An unsupervised method-oriented activation map is implemented by learning from Class Activation Mapping (CAM) \cite{RN403}. Since no image labels can be used to compute the weight of different channels in the feature map for unsupervised learning, the weight of each channel is set to 1. It is reasonable because each channel has an equal contribution to the test process. In Fig. \ref{cam PRID}, proposed scheme with local-aware module focuses on more accurate and fine-grained image regions than the baseline approach. At the same time, the baseline method has a large response on the head part of the person images. In Fig. \ref{cam Duke}, proposed scheme with global-aware module gives broader perspectives than baseline and retains more accurate feature response.  

	\subsection{Computational Complexity} \label{computationalcomplexity}
	This section first compares the FLOPs and trainable parameters between proposed uPMnet and other state-of-the-art methods. For different benchmarks, unsupervised methods usually have different trainable parameters. In this paper, the largest benchmark, DukeMTMC-VideoReID, is employed to make this comparison between proposed uPMnet and other state-of-the-art methods, as illustrated in Tabel \ref{ComputationalonDuke}. The proposed uPMnet using MobileNet as backbone only requires 1.8GFLOPs to run a single instance and achieves the best Rank-1 score. Besides, the FLOPs metric is highly related to the backbone. Benefiting from the effective MobileNet, the proposed uPMnet achieves the best performance with low computational complexity. 
	
	\begin{table}[!h]
		\centering
		\begin{threeparttable}
			\caption{Computational Complexity Comparisons on DukeMTMC-VideoReID.}\label{ComputationalonDuke}
			\begin{tabular}{lllll}
				\hline
				\multirow{2}{*}{Method} & \multirow{2}{*}{Backbone}  & \multicolumn{3}{c}{DukeMTMC-VideoReID} \\
				\cmidrule(r){3-5} 
				& & FLOPs      &  Params & Rank-1 \\ \hline
				TAUDL \cite{RN369} & ResNet50   & 5.3G  & 34.2M& 61.5*\\
				DAL \cite{RN87}    & ResNet50   & 4.7G  & 28.0M& 74.4*\\
				DAL \cite{RN87}    & MobileNet  & 0.7G  & 5.45M& 79.3*\\
				BUC \cite{RN247}   & ResNet50   & 5.3G  & 28.3M& 69.2 \\
				DBC \cite{RN118}   & ResNet50   & 5.3G  & 28.3M& 75.2 \\ \hline
				Proposed uPMnet    & ResNet50   & 10.0G & 93.3M& 81.3 \\
				Proposed uPMnet    & MobileNet  & 1.8G  & 32.3M& 83.6 \\ \hline
			\end{tabular}
			\begin{tablenotes}
				\item[1] `*' : the results are produced by us.
			\end{tablenotes}
		\end{threeparttable}
	\end{table}
	
	An ablation study for computational complexity is also represented, as illustrated in Tabel \ref{Computational3}. The benchmark PRID2011, iLIDS-VID, and DukeMTMC-VideoReID have 178, 300, and 2196 tracklets, respectively. For different benchmarks, the proposed uPMnet needs to initial intra-camera anchors $\mathcal{I}$ and cross-camera anchors $\mathcal{C}$ of different numbers. In other words, the proposed scheme needs different trainable parameters for different benchmarks. Besides, the computational complexity is also affected by the network architecture of different aware modules. Table \ref{Computational3} illustrates the computational complexity changes when using different aware modules on different benchmarks. The FLOPs and trainable parameters of the proposed uPMnet are the sum value from the local-aware network and global-aware network. Compared to the baseline, the proposed scheme using a local-aware or global-aware module achieves an average 2.6\% FLOPs increase along with an average 4.7\% Rank-1 increase for these three benchmarks. Although the proposed uPMnet, which fuses the local-aware and global-aware feature, increases FLOPs by an average value of 105.1\%, it improves the Rank-1 score with an average value of 11.4\%.

	\section{Conclusion}\label{conlusion}
	In this work, a robust part models-based scheme for unsupervised video reID is proposed. The proposed scheme is general and flexible and fuses part models and unsupervised learning decently. A local-aware module is employed to explore the potentials of part models, and a global-aware module is designed to overcome the disadvantages of part models. Features from local-aware module and global-aware module are fused to generate a rich and robust feature representation, which absorbs the advantages of part models without suffering from its disadvantages. Compared to the baseline method, proposed scheme achieves performance improvements significantly on each metric. Extensive evaluations show proposed approach achieves state-of-the-art performance on three video-based reID benchmarks. Concretely, compared to other state-of-the-art methods, proposed uPMnet achieves Rank-1 improvement of 6.7\%, 5.8\%, and 0.8\% on PRID2011, iLIDS-VID, and DukeMTMC-VideoReID benchmarks, respectively. Ablation study, parameter analysis, and qualitative analysis also verify the effectiveness of proposed scheme, local-aware module and global-aware module. 
	
	\normalem
	\bibliographystyle{iet2020}
	\bibliography{ref}
	
\end{document}